\documentclass[a4paper, 10pt]{article}
\usepackage[a4paper, margin=2cm]{geometry}  

\usepackage{graphicx}
\usepackage{subcaption} 

\usepackage[utf8]{inputenc}
\usepackage[T1]{fontenc}
\usepackage{lineno}
\usepackage{makecell}
\usepackage{tabularx}
\usepackage{pgfplots}
\usepackage{graphicx} 
\usepackage{pgf-pie} 
\usepackage{tikz}
\usepackage{svg}
\usepackage{graphicx}
\usepackage{hyperref}
\newcommand{\JournalTitle}[1]{#1}
\usepackage{amssymb}
\usepackage{changepage} 

\usepackage{pgfplots}

\usepackage{caption}

\usepgfplotslibrary{statistics}
\pgfplotsset{compat=1.16} 

\usepackage{graphicx}
\usepackage{subcaption} 

\title{DIPSER: A Dataset for In-Person Student Engagement Recognition in the Wild}
\author{
    Luis Marquez-Carpintero\textsuperscript{1} \and 
    Sergio Suescun-Ferrandiz\textsuperscript{1} \and 
    Carolina Lorenzo Álvarez\textsuperscript{2} \and
    Jorge Fernandez-Herrero\textsuperscript{2} \and
    Diego Viejo\textsuperscript{1} \and
    Rosabel Roig-Vila\textsuperscript{2} \and
    Miguel Cazorla\textsuperscript{1}
}
\date{
    \textsuperscript{1}Institute for Computer Research, P.O. Box 99. 03080, Alicante, Spain. Correspondence and requests for materials should be addressed to M.C (miguel.cazorla@ua.es) \\
    \textsuperscript{2}Faculty of Education, University of Alicante, 03690 Alicante, Spain; Correspondence to R.R (rosabel.roig@ua.es) \\
    \today
}
\begin{document}
\maketitle

\begin{abstract}
In this paper, a novel dataset is introduced, designed to assess student attention within in-person classroom settings. This dataset encompasses RGB camera data, featuring multiple cameras per student to capture both posture and facial expressions, in addition to smartwatch sensor data for each individual. This dataset allows machine learning algorithms to be trained to predict attention and correlate it with emotion. A comprehensive suite of attention and emotion labels for each student is provided, generated through self-reporting as well as evaluations by four different experts.

Our dataset uniquely combines facial and environmental camera data, smartwatch metrics, and includes underrepresented ethnicities in similar datasets, all within in-the-wild, in-person settings, making it the most comprehensive dataset of its kind currently available.

The dataset presented offers an extensive and diverse collection of data pertaining to student interactions across different educational contexts, augmented with additional metadata from other tools. This initiative addresses existing deficiencies by offering a valuable resource for the analysis of student attention and emotion in face-to-face lessons.

\end{abstract}

\section*{Background \& Summary}

The utilization of deep learning algorithms underscores the necessity for high-quality data to ensure appropriate generalization. This requirement becomes particularly pronounced when examining image datasets designed to assess student attention in the wild settings. In this context, the prevailing practice has been to collect data through online sessions \ref{table:dataset_description}, likely attributable to the ease associated with dataset recording. However, these datasets, often with recording sessions shorter than the 5 minutes analyzed here, often lack the integration of heart rate sensors, gyroscopes and accelerometers, components that the present dataset incorporates to investigate the correlation between these metrics.

In the course of our research, we have assembled a summary of the most prominent and pertinent datasets, as delineated in Table \ref{table:dataset_description}. This encompasses a detailed examination of seven datasets, juxtaposing them with our endeavor, accompanied by concise discussions on each. This comparative analysis serves not only to underscore the distinctive attributes of our dataset but also to contextualize our contribution within the expansive domain of attention measurement utilizing image datasets.

None of the datasets presented in Table \ref{table:dataset_description} have been recorded in an authentic classroom setting in the wild, nor have they utilized IMU data to accompany attention labels.

DAISEE\cite{gupta2016daisee}, a dataset providing RGB videos with a large number of subjects, is widely utilized. Despite its extensive adoption, several key issues have been identified in the application of this dataset for online classes. Predominantly featuring subjects of Indian origin (Malay Ethnicity), it exhibits limited variability for use in contexts with Western features. Moreover, the non-presential nature hampers observation of student interactions, and the excessively short duration of video clips further limits its applicability.

BAUM-1\cite{zhalehpour2016baum}, a dataset that includes Caucasian individuals and concentrates on the depiction of emotions, has its origins in a laboratory setting. All participants are actors, thereby potentially exaggerating their emotions for the camera. While connections between emotions and attention have been documented, the artificial setting of this dataset may detract from its utility in naturalistic studies.

The EmotiW dataset\cite{kaur2018prediction}, held in high regard within the domain of emotion and attention analysis, is distinguished for its emphasis on emotion and attention recognition in videos and its efficacy in assessing attention. Predominantly utilized in the annual Emotion Recognition in the Wild (EmotiW) Challenge, it encounters considerable limitations owing to its composition, which predominantly features subjects of Asian descent for online learning scenarios, thereby raising concerns regarding its generalizability across varied demographic landscapes.

\begin{figure}[!htb]
  \centering
  \subfloat[Subject number 4 from group 2]{\includegraphics[width=0.45\textwidth]{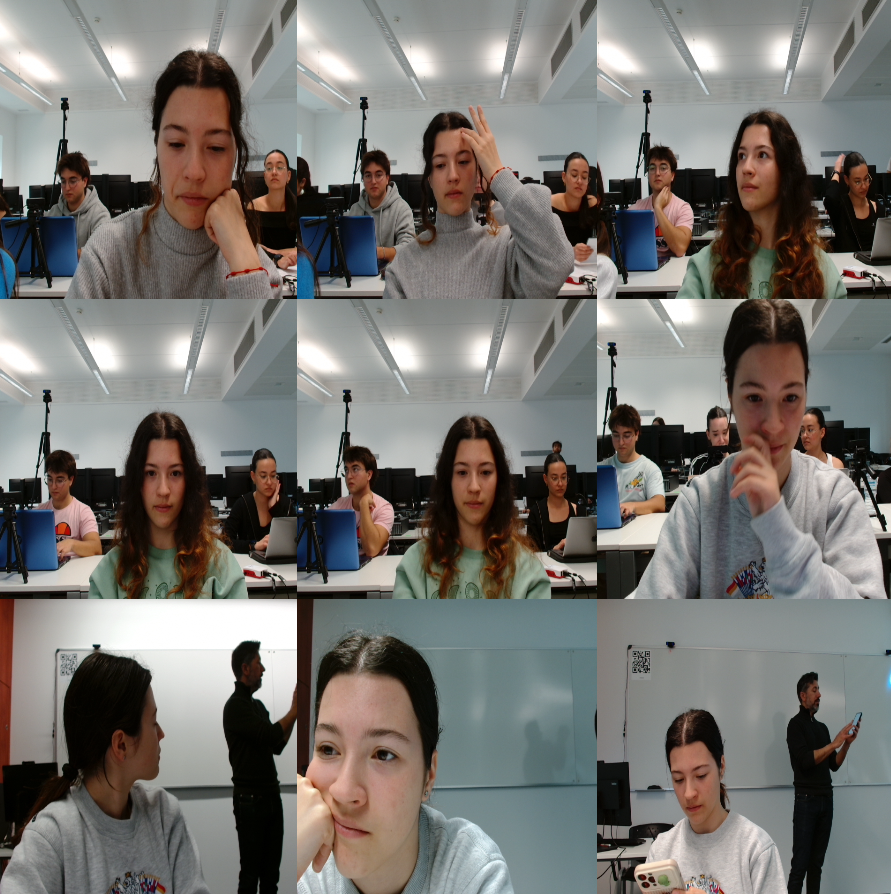}\label{fig:cara_chica}}
  \hfill 
  \subfloat[Subject number 19 from group 2]{\includegraphics[width=0.45\textwidth]{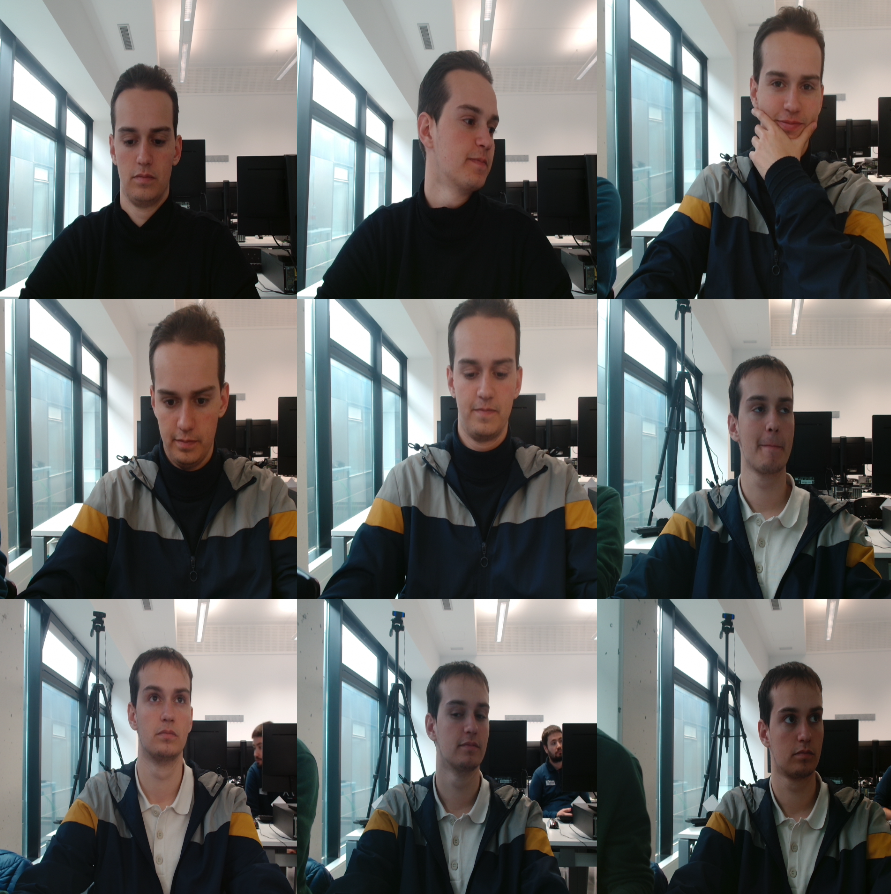}\label{fig:cara_chico}}
  \caption{The individual images of two subjects for each of the educational environments.}
  \label{fig:faces}
\end{figure}

The dataset introduced in this paper, denoted as DIPSER, Figure \ref{fig:faces}, offers an innovative method for quantifying attentional engagement in educational settings. It uniquely encompasses in-person sessions with Caucasian subjects and takes place in forested environments, targeting both undergraduate early childhood education (specifically, groups 1 and 2) and master's degree students (group 3). This methodology incorporates a wide range of scenarios and utilizes video recordings of five minutes in length, thereby providing a more comprehensive and versatile tool for attentional engagement studies.

The dataset provides high-resolution images at 640 × 480 pixels from individual cameras and 1280 × 720 pixels from the general cameras. This offers a contextual view of the students, capturing their posture and surroundings.

\begin{figure}[h!]
    \centering
    \includegraphics[width=\textwidth]{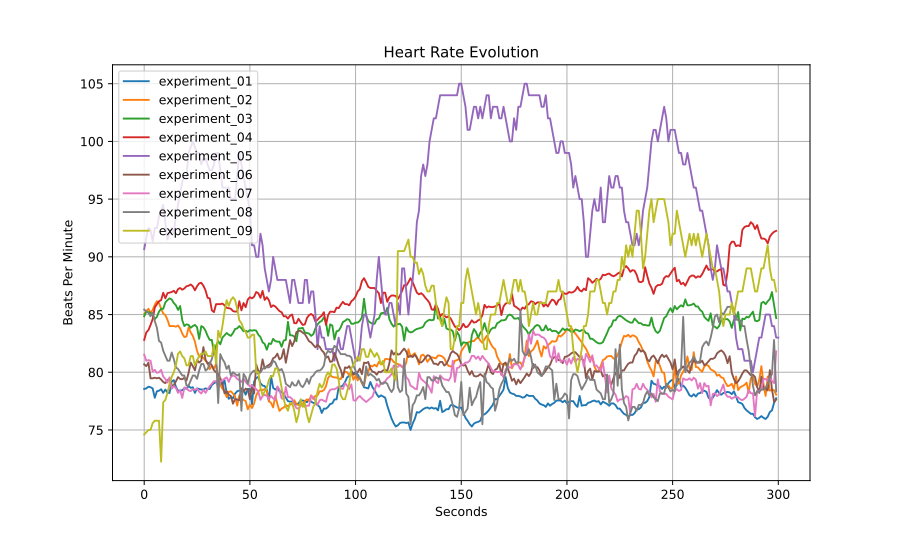}
    \caption{Changes in heart rate during the experiments (group 1)}
    \label{fig:heart_rate}
\end{figure}

A notable characteristic of DIPSER resides in its extensive scale, comprising precisely 1,311,761 images within the dataset. This voluminous collection translates to 3080.73 minutes of recording, devoid of any form of compression or encoding. This substantial volume of data ensures the dataset's suitability for training deep learning networks. It includes emotion and attention labels, along with smartwatch sensor data, such as heart rate (see Figure \ref{fig:heart_rate} for Group 01),  accelerometer, and gyroscope, recorded at every second of the sequences.
These characteristics endow the dataset with a high degree of variability, challenging the generalization capabilities of algorithms (refer to the Technical Validation section for more details).

Furthermore, the dataset includes processed image files featuring bounding boxes for faces and bodies, along with annotations for age, gender, face mesh, hand mesh, body mesh, gaze estimation, and head pose. These comprehensive annotations serve to facilitate the work of other researchers, enhancing accessibility and usability.


\section*{Methods}

This section outlines the methodology employed in the development and validation of the DIPSER dataset. The objective of this dataset is to capture and analyze in-person student engagement in naturalistic classroom settings using a combination of video and sensor data. The following subsections provide a detailed account of the experimental design of the sessions, data collection setup, synchronization processes, data acquisition protocols, and techniques used for data processing and labeling.

\subsection*{Overview}
The proposed dataset is designed for use with various objectives, such as tracking the evolution of emotions during a class according to the scale of Pekrun et al.\cite{pekrun2007control, pekrun2011measuring} or their co-dependence with attention according to the scale of Goldberg et al.\cite{goldberg2021attentive}.

The methodology employed in this study is a quasi-experimental design, as three pre-existing groups of students were used without random assignment, and no control group was included since all interventions were applied to all individuals. The interventions consisted of nine experimental activities or sessions, or experiments, employing different didactic methodologies. This design was chosen as an initial approach to the research subject, following recommendations such as those by Ungar and Kording \cite{liu2021quantifying}, particularly in combination with the application of machine learning (ML) for data enhancement, which is relevant in this case. Additionally, this type of methodological design has been observed in several studies on artificial intelligence tools in the educational field \cite{ouyang2023integration, chen2022use}.

For each subject, the dataset includes Inertial Measurement Unit (IMU) data, RGB images, global context cameras, and additional post-processed data from supplementary tools. The dataset comprises an RGB dataset featuring nine distinct experiments, involving a varying number of students—up to 20 for groups 1 and 2, and up to 16 for group 3. These experiments span the following variety of educational settings:

\begin{itemize}
\item \textbf{Session 1: News Reading} - Students read news articles either projected on a screen or on their personal devices, focusing on content that will be assessed later.
\item \textbf{Session 2: Brainstorming Session} - During this creative session, students generate ideas for projects or solutions to problems.
\item \textbf{Session 3: Lecture} - A traditional teaching format where the instructor delivers a lecture in front of the class, with minimal to no student interaction.
\item \textbf{Session 4: Information Organization} - Students organize and synthesize information gathered from various sources.
\item \textbf{Session 5: Lecture Test} - A formal assessment concerning the content of Session 3, administered via mobile devices.
\item \textbf{Session 6: Individual Presentation of Work} - Students randomly selected present their projects to the group.
\item \textbf{Session 7: Knowledge Test} - A formal written assessment on a specific subject area, conducted using Kahoot to evaluate collective knowledge on the material covered in Session 1.
\item \textbf{Session 8: Robotics Experimentation} - A practical session where students apply robotics technology to problem-solving, emphasizing computational thinking.
\item \textbf{Session 9: MTINY Activity Design} - Students design and plan an educational activity using the MTINY educational tool, integrating computational thinking principles.
\end{itemize}

\subsection*{Capture Setup}

\begin{figure}[!htb]
    \centering
    \includegraphics[width=1\textwidth]{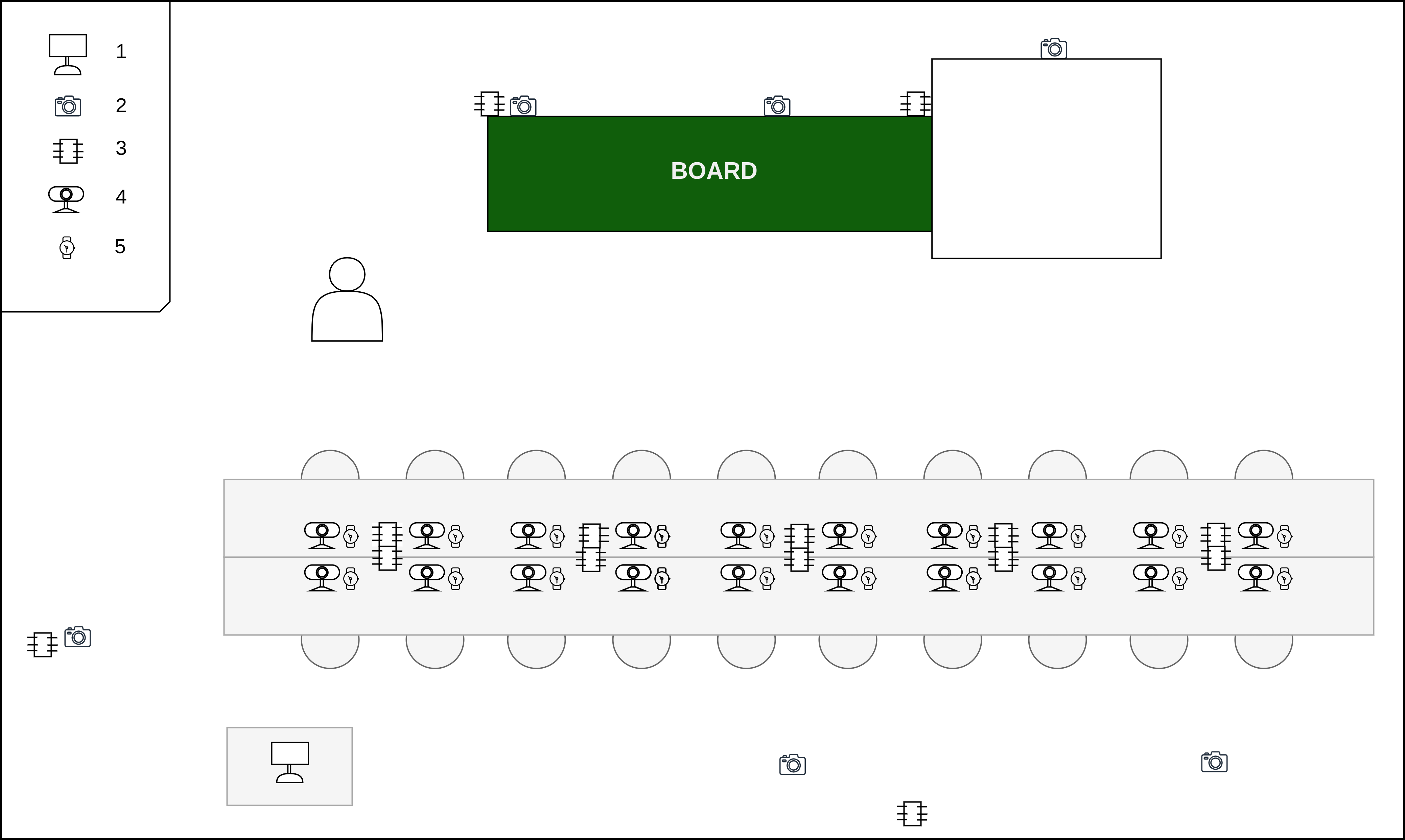}
    \caption{Capture setup for experiments 6 to 9.}
    \label{fig:setup2}
\end{figure}

As shown in Figures \ref{fig:setup1} and \ref{fig:setup2}, the recording setup was implemented in a standard classroom at the University of Alicante within the Faculty of Education, as extensively detailed in other papers \cite{marquez2023author}. Up to six general context cameras were installed, capable of recording multiple students simultaneously. Additionally, a smartwatch and a camera were placed at each student's desk to capture the micro-gestures of the participants.

The processing of the camera and smartwatch data was centralized on Raspberry Pi Model 4-B units. Despite of being low-cost devices, each Raspberry Pi was able to handle two cameras and smartwatches simultaneously.

For each setup, the cameras were connected to the USB 3.0 port of a Raspberry Pi Model 4-B. The smartwatches transmitted the sensor-collected information to the Raspberry Pi, which was responsible for storing both the images and all content received from the watches via a web server.

For the Wi-Fi connection, a router without internet access was used specifically to connect the Raspberry Pis and smartwatches to the same network.

In the case of context cameras, each pair also required a Raspberry Pi. However, in such cases, the resolution and data capture frequency varied. Due to subjects being further away, it was preferred to increase the resolution and decrease the FPS.

Finally, for experiments requiring collaboration among participants and a redistribution of tables, the entire camera setup was maintained, Figure \ref{fig:setup2}, and each camera was kept at the same distance from the users as in other experiments. This setup occasionally resulted in users moving out of frame when leaning over to speak with their peers.

\subsection*{Procedure}

This section outlines the procedure followed to record the dataset, divided into the following subsections:

Initially, the Raspberry Pis and smartwatches were turned on, ensuring they were connected to the same network. Then, from a computer acting as a server for each Raspberry Pi (a computer dedicated to this purpose), a signal was sent to simultaneously delete all stored images to prevent memory overflow. Subsequently, it was confirmed that the order of cameras A and B matched the assigned physical location. Finally, the web server of each Raspberry Pi was started using a remote command from the main computer.

At this point, the Raspberry Pis were ready to begin recording. However, first, the IP address and the camera associated with each subject's smartwatch had to be configured. This ensured that the individual camera and the watch were targeting the same subject.

Once this was done for all students, the recording of the images (both individual and general) commenced.

To terminate the experiment, a new command was issued. Given that the cessation of recording was not precisely simultaneous—owing to a slight phase delay—a recording buffer was established both before and after the experiment. This buffer allowed for the subsequent processing of the data and its trimming to the precise duration of the experiment.

\subsection*{Data Acquisition}

As mentioned, data acquisition was performed by centralizing commands and using embedded devices for every two cameras, with an additional smartwatch connected to the Raspberry Pi for each subject. After completing each recording session, the images were transferred from the Raspberry Pis to a hard drive. This allowed for the deletion of the previous day's content and ensured sufficient storage space for the new recording session.




\section*{Data Records}

We present a structured dataset with the following directory organization for scientific research.
This section reviews different aspects of the data records we provide. We focus on the technical features of each one. 
More information about how the data is organized can be found at the Code Availability.

\subsection*{General or Contextual Images}

\begin{figure}[!htb]
    \centering
    \includegraphics[width=0.6\textwidth]{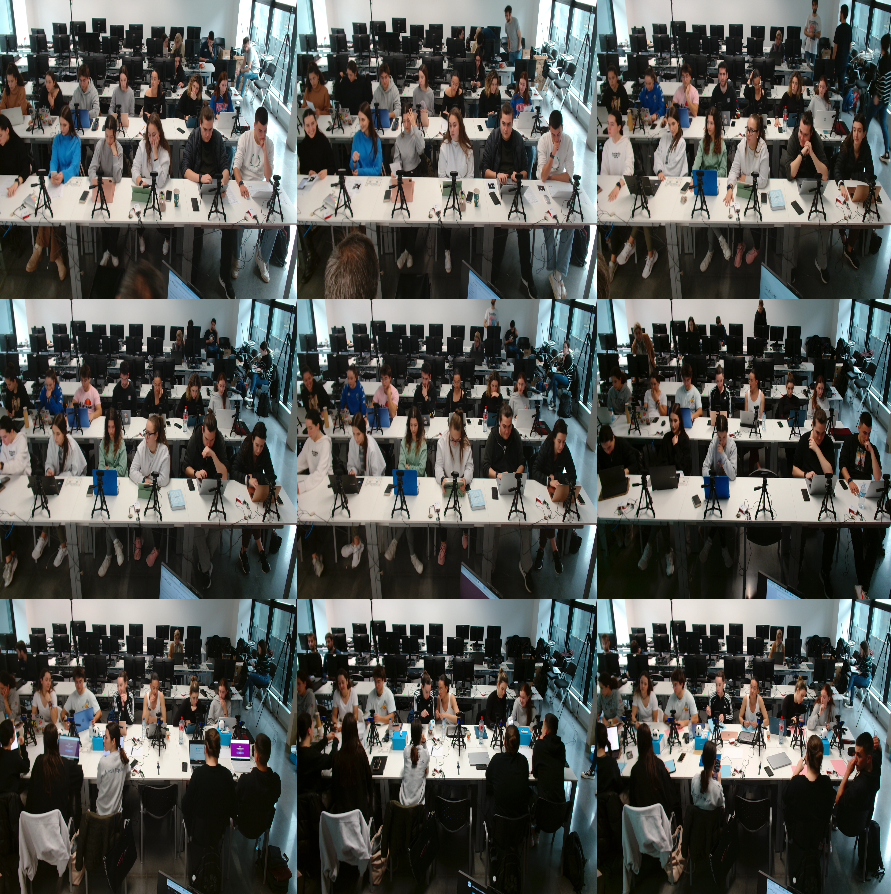}
    \caption{General image of a context camera for each learning environment.}
    \label{fig:general_cam}
\end{figure}

In each conducted experiment, comprehensive images capturing the overall context, shown in Figure \ref{fig:general_cam}, which include various participants, were collected from distant vantage points for every group involved. These images were captured from heights approximately ranging between 1.8 and 2.3 meters, utilizing an array of five or six cameras, contingent upon the specific experimental requirements. This arrangement was designed to ensure exhaustive coverage of the area, thereby effectively mitigating potential blind spots. The cameras recorded RGB images at a frame rate of approximately 9 fps. Each image was captured with a resolution of 1280x720 pixels, providing detailed spatial information for subsequent analysis.

\subsection*{Individual Images}
For every participant in the study, individual RGB images focusing on the subjects' faces were captured. This specific approach facilitated the collection of detailed facial expressions and features, which are essential for studies on emotion and attention. In this context, the speed of image capture was prioritized above resolution to adequately capture microfacial expressions. The images were recorded at a frame rate of approximately 10 fps, ensuring temporal consistency across various datasets, and were captured with a resolution of 640x480 pixels, offering a good definition detail for each subject.
    
\subsection*{Sensor Data}
This manuscript presents sensor data derived from various instruments, notably including the Linear Acceleration Sensor, Gyroscope, Light Sensor, Rotation Vector, and Cardiac Frequency Sensor, structured in a JSON file format. The measurements obtained from these sensors are quantified in standardized units, ensuring standardization and comparability across datasets. The configuration shown in Table \ref{table:sensor_data}  allows for the assembly of temporally dense datasets. Data points are consolidated at one-second intervals, promoting a systematic framework for data storage and retrieval. Consequently, to pinpoint a specific sensor reading, researchers are required to access the file corresponding to the second immediately following the desired measurement point.

\subsection*{Setup Configuration}
\begin{figure}[!htb]
    \centering
    \includegraphics[width=1\textwidth]{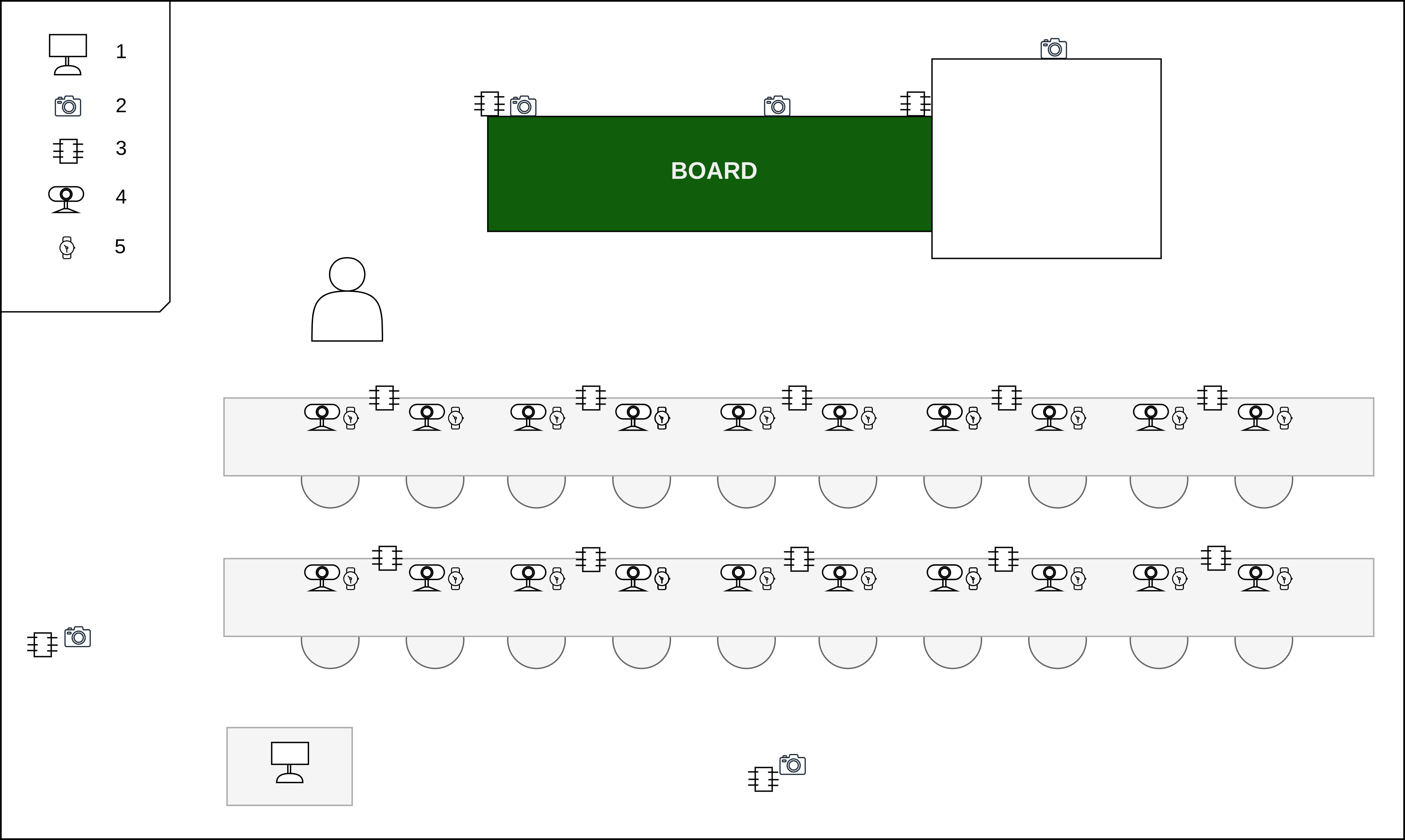}
    \caption{Capture setup for experiments 1 to 5.}
    \label{fig:setup1}
\end{figure}

As previously mentioned and illustrated in Figure \ref{fig:setup1} and Figure \ref{fig:setup2}, a total of 20 individual cameras, indicated in the diagram as number 4, and clocks, marked as number 5, have been positioned according to the arrangement of the students during the recording session. Due to the technical limitations of the embedded devices, it is necessary to place one of these devices, identified as number 3, for every pair of cameras.

In the same figures, it is highlighted how the cameras intended for recording the context, marked with the number 2, also require, at a minimum, one embedded device for every two units. Finally, Figure \ref{fig:setup1} shows the personal computer designated to function as a server, labeled with the number 1. This server is intended to monitor and execute commands in a centralized manner to all embedded device units, in addition to handling the synchronization of time among the embedded devices, which, in turn, will synchronize their time with each of the smartwatches.

In the setup described, up to 26 Logitech HD Pro C920 cameras have been used, each featuring a 78-degree field of view. These cameras are associated with items labeled as number 2 and number 4 in the diagram. Additionally, 20 Samsung Galaxy Watch 5 smartwatches are utilized, one for each student, identified in the diagram as number 5. A Lenovo IdeaPad 5 Pro Gen 6 laptop has been chosen to serve as the server, corresponding to number 1 in the diagram. Lastly, the embedded devices are Raspberry Pi Model 4-B with 2GB of RAM.

\subsection*{Labeling}
Labels are meticulously provided for each subject, with a dedicated file indicating the frames, levels of attention \cite{goldberg2021attentive}, and emotions assessed \cite{pekrun2011measuring}. Attention is measured on a scale from one to five, with 5 indicating maximum attention. Emotion is categorized into nine categories: 9: Enjoyment, 8: Hope, 7: Pride, 6: Relief, 5: Anger, 4: Anxiety, 3: Shame, 2: Despair, 1: Boredom.

Additionally, proprietary software specifically tailored for this purpose has been developed. A total of five evaluators per student participate in this process, consisting of four experts and one self-evaluation, to ensure a comprehensive and balanced assessment. The fourth and fifth evaluators were supplemented to ensure these five tags per student.

Self-evaluation is conducted by students after each recording session, where they proceed to label the previous day's session. Subsequently, students are contacted again in cases where:
\begin{enumerate}
    \item the student did not attend and the session was not labeled, \item the labels do not meet the minimum threshold of three combined changes in emotion or attention per experiment. In this case, labeling is repeated until this threshold is met.
\end{enumerate}

The labeled frames indicate changes in attention or emotion at the specified second, and evaluators are presented with only one frame per second.

\subsection*{Processing with Additional Tools}
To facilitate the analysis and processing of the collected data, images of each participant undergo preprocessing with specialized algorithms. The results are meticulously recorded in a JSON file, which includes the rotation angle needed for horizontal eye alignment, determined using the Opal \cite{cobo2024representation} tool; bounding boxes for the face and body, identified with the YOLO algorithm via the MiVOLO\cite{kuprashevich2023mivolo} tool; as well as demographic details like the participant's age and gender, obtained also using the MiVOLO tool. Additionally, landmarks for the hands\cite{zhang2020mediapipe} and the facemesh\cite{grishchenko2020attention} are generated using MediaPipe. For the pose estimation we use Mediapipe tool that use BlazePose\cite{bazarevsky2020blazepose}.
Finally, we employ DeepFace \cite{serengil2021lightface} to obtain ethnicity and emotion data. This tool provides a comprehensive dataset for advanced analysis of morphology and gestures. Although the ethnicity determined by this tool does not correspond to that assigned by \cite{belen2018cranial}, given its definition, Caucasian ethnicity is considered to encompass the White ethnicity.

\subsection*{Organization}
The dataset is meticulously organized into three primary groups, each corresponding to different student cohorts involved in the classroom recordings. Within each cohort, the dataset is further subdivided into nine distinct experiments or scenarios, each scenario comprising individual folders for each subject. The directory structure is designed for intuitive navigation and efficient data access, adhering to a consistent format:
\texttt{/group1/experiment2/subject15.zip}

Within each subject's compressed \texttt{.zip} file, the data is sorted into specific subfolders: \texttt{images}, \texttt{watch\_sensors}, \texttt{labels}, and \texttt{metadata}. Each subfolder is dedicated to a particular type of information, thus ensuring a high level of organization and ease of access for researchers engaging with the dataset.

\section*{Technical Validation}

This section presents a comprehensive qualitative and quantitative analysis to confirm the reliability of the data set.

Emphasizing the examination of the dataset, we conducted a thorough review, analyzing the population data provided by a model rather than official records due to privacy concerns. This approach facilitated an inclusive analysis, taking into account diverse ages and gender variations, thus ensuring a multifaceted understanding of the dataset's applicability and robustness.

\subsection*{Quality and Reliability}

\begin{enumerate}
    \item \textbf{Camera and Clock Synchronization:} To synchronize the cameras and clocks, we utilized a software-based multi-device synchronization system that requests the time from a central server. Subsequently, we verified the synchronization through forced brightness changes in each experiment to confirm accurate synchronization, ensuring that, prior to the dataset's publication, the discrepancy between the clock timestamp and the camera's timestamp is less than 0.5 seconds.
        
    \item \textbf{Completeness:} We verified that the dataset extensively encompasses the intended domain, capturing all pertinent attributes, such as student attention levels, emotional states, and associated biometric indicators. This ensures a holistic view of the data's applicability to real-world scenarios.
    
    \item \textbf{Consistency:} Data collection methodologies were examined for uniformity across different sessions and environmentsFigure \ref{fig:setup1} and Figure \ref{fig:setup2}. Consistent methods are vital for the dataset's reliability, ensuring that variations in data are genuine and not artifacts of the collection process. No significant data absence was detected beyond random variability due to the performance of embedded devices.
    
    \item \textbf{Temporal Relevance:} The dataset's currency was assessed, including the presence of timestamps where necessary. This enables the analysis of temporal trends and the dataset's relevance to contemporary studies.
    
\end{enumerate}

\subsection*{Integrity and Data Structure}
\begin{figure}[h!]
    \centering
    \includegraphics[width=0.75\textwidth]{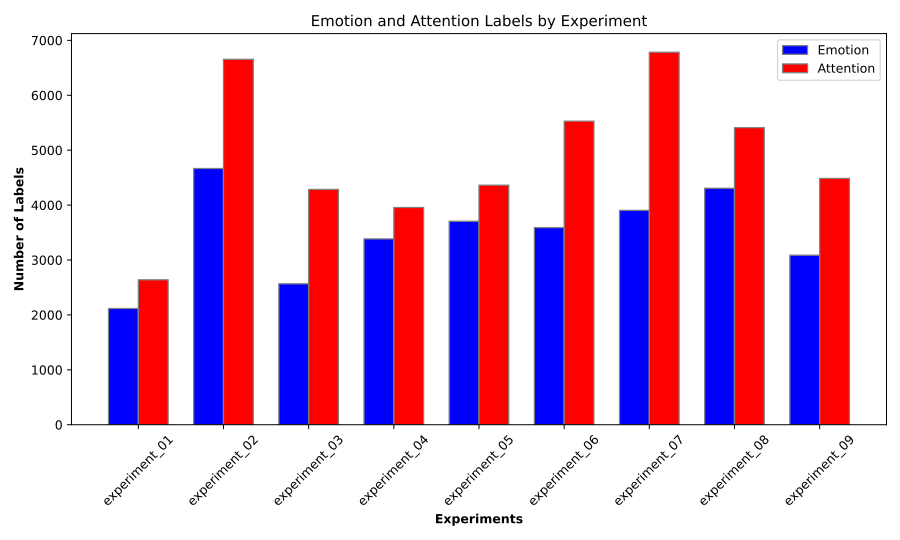}
    \caption{Number of labels per experiment}
    \label{fig:number_labels}
\end{figure}

\begin{enumerate}
    \item \textbf{Data Structure:} The organization of the dataset, including its schema and formatting, was scrutinized to confirm its efficacy in supporting efficient data retrieval and analysis. A well-structured dataset is foundational for robust research outcomes.
    \item \textbf{Data Consistency:} The consistency of the dataset was thoroughly examined to ensure uniformity across different sessions and environments. Consistent methods are vital for the dataset’s reliability, ensuring that variations in data are genuine and not artifacts of the collection process.
    \item \textbf{Data Types and Formats:} It was verified that each column in the dataset employs the appropriate data type and format, thereby facilitating accurate and efficient analysis. Furthermore, the willingness to have a substantial number of labels on each subject was evaluated by obtaining those experiments most susceptible to changes in emotion and attention, Figure \ref{fig:number_labels}.
\end{enumerate}

\section*{Future Work}
As part of our ongoing efforts to enrich the dataset's diversity, future work will focus on expanding the heterogeneity of the student body. This will include incorporating students from various ethnic backgrounds and academic levels, recognizing the diverse profiles that access different university degrees or master's programs. Such an expansion is anticipated to further enhance the dataset's generalization capabilities and its applicability to a broader spectrum of educational and psychological studies.


\section*{Usage Notes}

"DIPSER" presented in this study offers a comprehensive collection of multimodal data designed to support research in attention analysis, behavior understanding, and affective computing. To aid researchers in optimizing the use and impact of this dataset, we provide the following guidelines and considerations:

\subsection*{Software Recommendations}
\begin{itemize}
    \item \textbf{Image and Video Analysis:} For the processing and analysis of RGB images and videos, software packages such as OpenCV for Python, MATLAB's Image Processing Toolbox, and the Deep Learning Toolbox are recommended. These tools facilitate the implementation and application of machine learning models.
    \item \textbf{Sensor Data Analysis:} Tools like Pandas for Python are recommended for managing and analyzing time-series data efficiently. R packages, including dplyr and tidyr, are also suitable for data manipulation and analysis.
    \item \textbf{Facial and Emotion Recognition:} For specialized analyses, such as facial and emotion recognition, frameworks like TensorFlow or PyTorch are recommended. These offer pre-trained models and libraries to streamline development.
\end{itemize}

\subsection*{Ethical and Privacy Considerations}
Given the sensitive nature of attention data, which may include personal and biometric information, stringent privacy controls have been implemented:

\begin{itemize}
\item \textbf{Anonymization:} All identifiable information has been either removed or anonymized in adherence to relevant data protection regulations, with the exception of faces, which are necessary for processing attention and emotion using deep learning algorithms.
\item \textbf{Access Controls:} Dataset access is regulated, requiring applicants to outline the intended data use. Researchers must commit to adhering to ethical guidelines concerning data utilization.
\item \textbf{Use Restrictions:} The dataset is strictly for academic and research purposes. Any commercial application intentions must receive explicit approval from the data custodians.
\end{itemize}

The university's ethics committee has approved the following measures after a comprehensive review focused on privacy and consent matters, particularly given the sensitive nature of the RGB and IMU data related to student attention and engagement. Moreover, we adhere to the guidelines of the Declaration of Helsinki\cite{goodyear2007declaration} for research involving human subjects, as recommended by the World Medical Association (WMA) for application in other fields. This data will be used solely for academic purposes and not for commercial objectives.
The study was conducted in accordance with the Declaration of Helsinki \cite{goodyear2007declaration} and was approved by the Ethics Committee of the University of Alicante, where the students were recorded. All participants signed an informed consent previous to the experiments. The dataset has been made publicly available under the CC-BY license. These data will be utilized exclusively for academic purposes and will not be employed for commercial purposes.

\section*{Code Availability}
We have uploaded Python code on GitHub, available at https://bitbucket.org/rovitlib/dipser/, which is capable of reading labels at each time instance, in addition to sensor data and images. This code intelligently groups the data with context from cameras to efficiently display a pandas DataFrame, effectively labeling each data point for individual students. Any relevant information about the code can be found at the repository link.



\section*{Acknowledgements}

This project has been developed under the framework of the CIPROM/2021/17 Prometeo project entitled "Meebai: Una metodología para la educación consciente de las emociones basada en la inteligencia artificial".

\section*{Author contributions statement}


L.M. contributed to software development, experimental design, validation, data processing, manuscript drafting, and data analysis. S.S. participated in the experimental design and software development. C.L., J.F. were responsible for designing class dynamics and conducting expert labeling. D.V. provided leadership in project direction and supervision and made significant contributions to the manuscript's writing and review. R.V. and M.C. led the project, managed supervision, facilitated the acquisition of funding for the experiment, provided resources, and actively participated in the manuscript's writing and review process. All authors reviewed the manuscript.

\section*{Competing interests}

Te authors declare no competing interests.

\newpage

\section*{Figures \& Tables}





\begin{figure}[h!]
    \centering
    \begin{subfigure}[b]{0.45\textwidth}
        \centering
        \includegraphics[width=\textwidth]{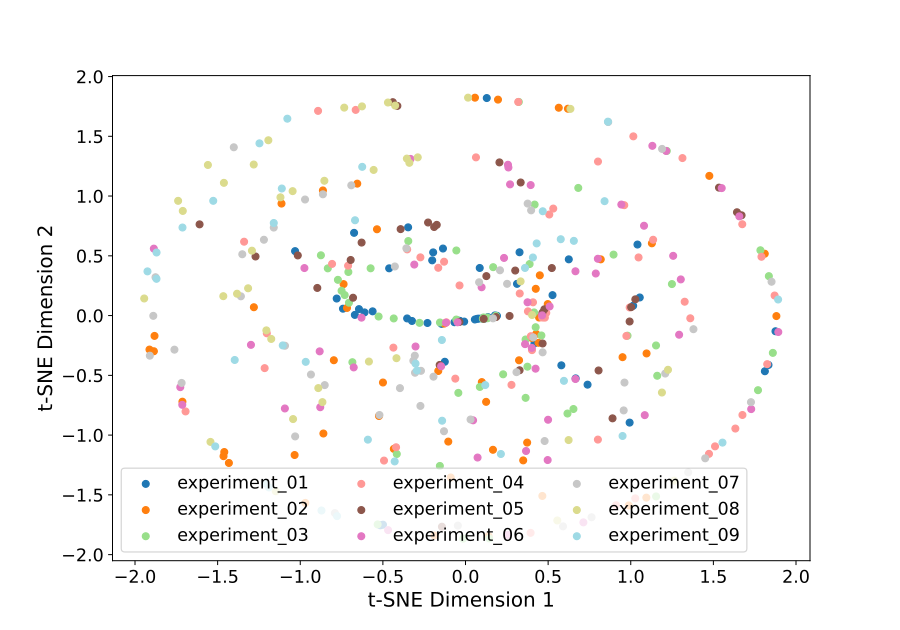}
        \caption{LSM6DSO Gyroscope Sensor}
        \label{fig:plot_inst_hr1}
    \end{subfigure}
    \hfill
    \begin{subfigure}[b]{0.45\textwidth}
        \centering
        \includegraphics[width=\textwidth]{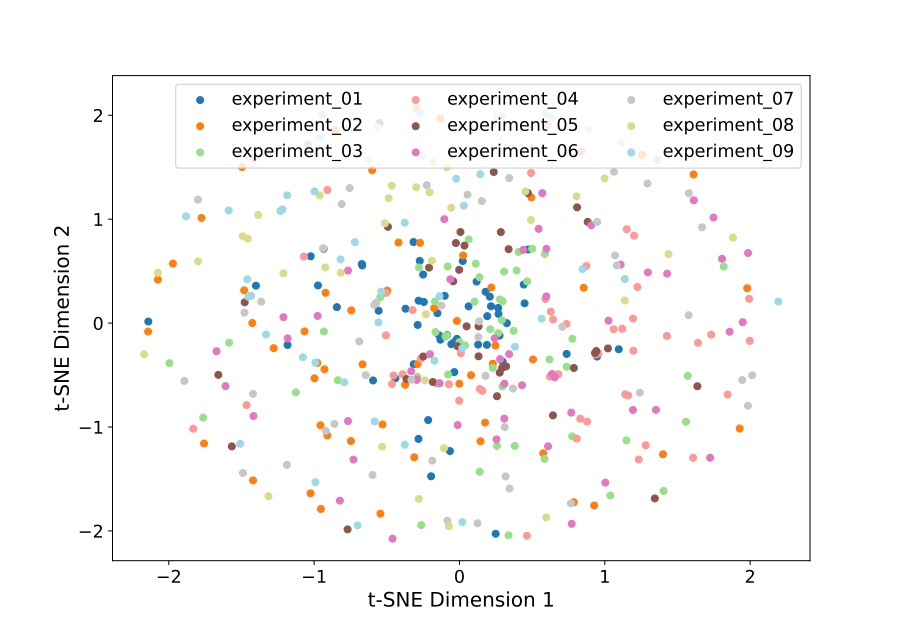}
        \caption{Samsung Linear Acceleration Sensor}
        \label{fig:plot_inst_hr2}
    \end{subfigure}
    \par\bigskip
    \begin{subfigure}[b]{0.45\textwidth}
        \centering
        \includegraphics[width=\textwidth]{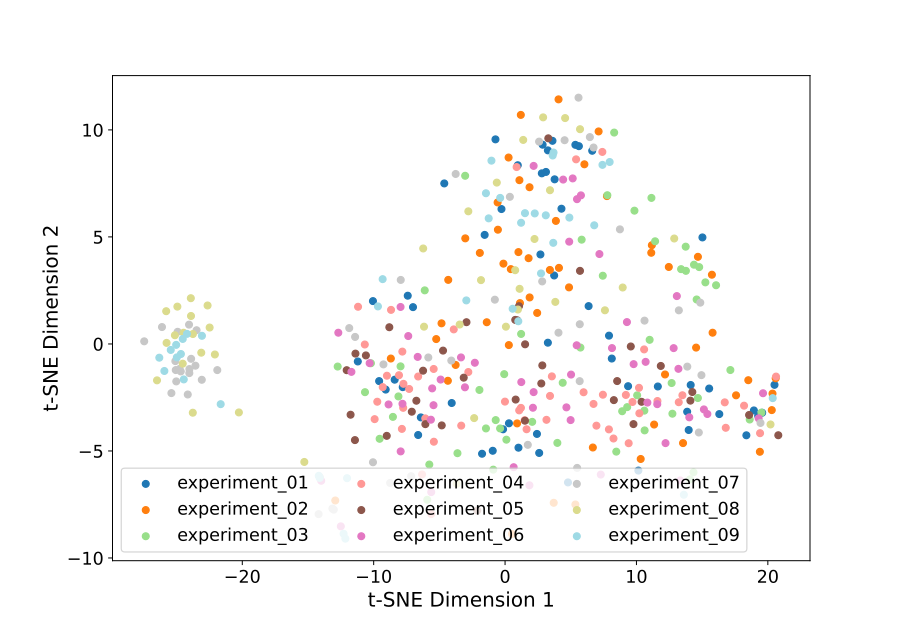}
        \caption{Samsung Rotation Vector}
        \label{fig:plot_inst_hr3}
    \end{subfigure}
    \hfill
    \par\bigskip

    \caption{Dimensional reduction TSNE watch sensors}
    \label{fig:plot_tsne}
\end{figure}

\begin{figure}[h!]
    \centering
    \begin{subfigure}[b]{0.45\textwidth}
        \centering
        \includegraphics[width=\textwidth]{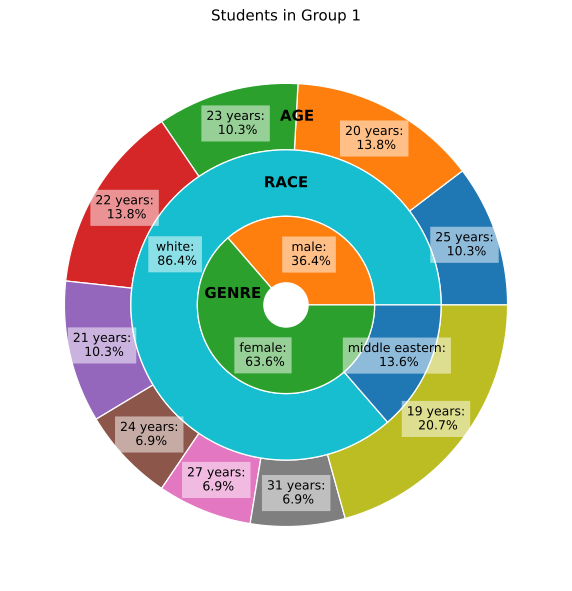}
        \caption{Descriptive statistics for group 1 participants}
        \label{fig:plot_inst_hr1}
    \end{subfigure}
    \hfill
    \begin{subfigure}[b]{0.45\textwidth}
        \centering
        \includegraphics[width=\textwidth]{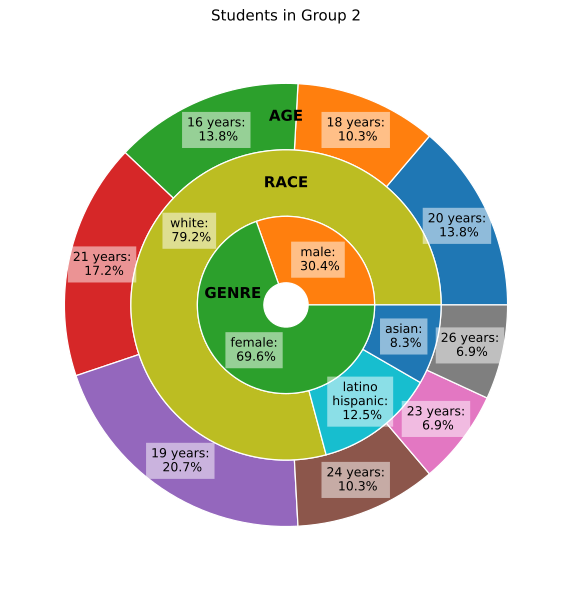}
        \caption{Descriptive statistics for group 2 participants}
        \label{fig:plot_inst_hr2}
    \end{subfigure}
    \par\bigskip
    
    \begin{subfigure}[b]{0.45\textwidth}
        \centering
        \includegraphics[width=\textwidth]{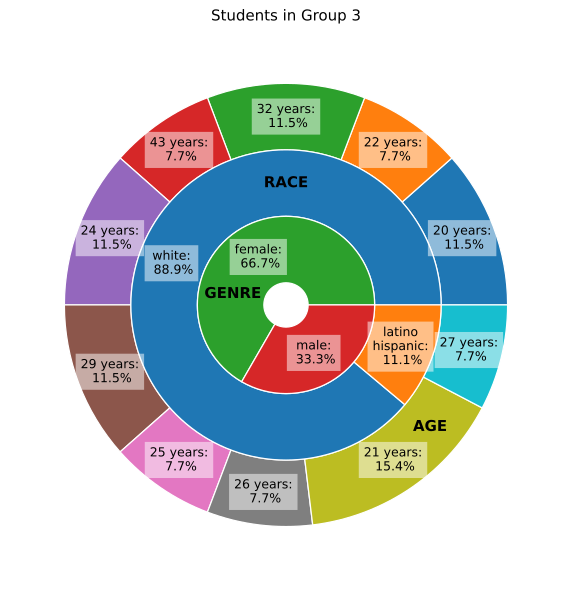}
        \caption{Descriptive statistics for group 3 participants}
        \label{fig:plot_inst_hr3}
    \end{subfigure}
    \hfill
    \par\bigskip

    \caption{Descriptive statistics of students per group}
    \label{fig:students_info}
\end{figure}

\begin{table}[!htb]

\begin{adjustwidth}{-4cm}{-4cm} 
\centering
\small 
\begin{tabular}{|l|l|l|l|l|l|l|l|l|}
\hline
\textbf{\makecell{Dataset}} & \textbf{\makecell{Predominant \\ Ethnicity}} & \textbf{\makecell{Hours}} & \textbf{\makecell{Scenarios}} & \textbf{\makecell{Subjects}} & \textbf{\makecell{Context\\Cameras}} & \textbf{\makecell{IMU}} & \textbf{\makecell{In the\\Wild}}& \textbf{\makecell{Labels}} \\ \hline

DAiSEE\cite{gupta2016daisee} & Malay & 25 & E-Learning & 112 & None & x & $\checkmark$ & \makecell[tl]{Engage: very
low, low,\\
 high, very high} \\ \hline

BAUM-1\cite{zhalehpour2016baum} & Malay & - & E-Learning & 31 & None & x & x &\makecell[tl]{ Happiness, Sadness, \\ Fear, Anger, Disgust, \\Confusion, Boredom, \\Interest}
 \\ \hline

EmotiW 2018\cite{kaur2018prediction} & Malay & 16.5 & E-Learning & 78 & None & x & $\checkmark$ & Engagement {0, 1, 2, 3} \\ \hline

EngageNet\cite{singh2023have} & Malay & 31 & E-Learning & 127 & None &  x & $\checkmark$ &\makecell[tl]{ Highly Engaged \\Engaged Barely \\Engaged and \\ Not Engaged} \\ \hline

NVIE\cite{wang2010natural} & Mongolian & - & E-Learning & 215 & None & x & $\checkmark$ & - \\ \hline

\makecell[tl]{Student Engagement\\Dataset\cite{Delgado_2021_ICCV}} & \makecell[tl]{Caucasian} & - & E-Learning & 19 & None  & x & $\checkmark$ & \makecell[tl]{Looking at their \\paper, Looking at \\their screen and \\Otherwise/Wandering} \\ \hline

HBCU Dataset\cite{whitehill2014faces} & Ethiopian & $\sim25.5$ & E-Learning &  34 & None & x & $\checkmark$ &  \makecell[tl]{Not engaged at all,\\ Nominally engaged,\\ Engaged in task,\\ Very engaged,\\ X: The clip/frame\\ was very unclear} \\ \hline

\textbf{DIPSER (OURS)} & Caucasian \ref{fig:students_info} & 51.3 & Face-to-Face & 54 & Until 6 & $\checkmark$ & $\checkmark$ & \makecell[tl]{Engagement: 5 levels \\ Emotion: 9 types} \\ 

\hline
\end{tabular}
\end{adjustwidth}
\caption{Summary of major RGB attention datasets Features: ethnic dominance \cite{belen2018cranial}, recording duration and environment, subject counts, additional context cameras, IMU presence, in-the-wild collection, and labeling tags.}
\label{table:dataset_description}
\end{table}

\begin{table}[!htb]
\centering
\resizebox{\textwidth}{!}{%
\begin{tabular}{|p{3cm}|p{5cm}|p{5cm}|p{2cm}|p{3.5cm}|}
\hline
\textbf{Official Sensor Name} & \textbf{Description} & \textbf{Measured Values} & \textbf{Samples per Second} & \textbf{Units in Data Sampling} \\
\hline
Samsung HR None Wakeup Sensor & Detects the heart rate & Beats Per Minute & 1 & Beats Per Minute \\
\hline
Samsung Linear Acceleration Sensor & Reports linear acceleration excluding gravity in the sensor frame. & 
Value 0: X-axis, Value 1: Y-axis, Value 2: Z-axis & 100 & 
Value 0: m/s\textsuperscript{2}, Value 1: m/s\textsuperscript{2}, Value 2: m/s\textsuperscript{2} \\
\hline
LSM6DSO Gyroscope & Reports the rate of rotation around three sensor axes. & 
Value 0: rad/s, Value 1: rad/s, Value 2: rad/s & 100 & 
Value 0: m/s\textsuperscript{2}, Value 1: m/s\textsuperscript{2}, Value 2: m/s\textsuperscript{2} \\
\hline
OPT3007 Light Sensor & Measures the level of environmental light. & Luminosity value & 5 & Lux units \\

 \hline
\end{tabular}}
\caption{\label{table:sensor_data}Description of various sensors with their respective data values.}
\end{table}


\end{document}